# Enhancing Video Understanding: Deep Neural Networks for Spatiotemporal Analysis

AmirHosein Fadaei

School of Electrical and Computer Engineering, College of Engineering, University of Tehran, Tehran, Iran;
hosein.fadaei@ut.ac.ir

Mohammad-Reza A.  Dehaqani [*]

School of Electrical and Computer Engineering, College of Engineering, University of Tehran, Tehran, Iran; School of Cognitive Sciences, Institute for Research in Fundamental Sciences (IPM), Tehran, Iran; dehaqani@ut.ac.ir

It's no secret that video has become the primary way we share information online. That's why there's been a surge in demand for algorithms that can analyze and understand video content. It's a trend going to continue as video continues to dominate the digital landscape. These algorithms will extract and classify related features from the video and will use them to describe the events and objects in the video. Deep neural networks have displayed encouraging outcomes in the realm of feature extraction and video description. This paper will explore the spatiotemporal features found in videos and recent advancements in deep neural networks in video understanding. We will review some of the main trends in video understanding models and their structural design, the main problems, and some offered solutions in this topic. We will also review and compare significant video understanding and action recognition datasets.

## 1 INTRODUCTION

According to Cisco predictions, in 2022, 82% of the web data consisted of videos. It will take 5 million years for someone to watch it all [26]. It is interesting to note the rise in popularity of using videos for transferring or presenting data on the web and various applications. Video is a more engaging and visually appealing way to convey information to users. The video consists of images, text, and sound and can easily show the relationships and how objects interact with each other with this multimodal structure. They are also much simpler for the user to understand.

The complete video provides more information than the individual frames presented on their own. With a video, you can track changes and how actions are taking place. The movements, changes in shape and color, the time and order of the different events, and all other patterns seen through time are partial examples of information only available with a video, and images can not contain them. This information proved useful and very important for classification and understanding the context at a higher cognition level and for more complex tasks. For example, it is shown with data that facial movements can help us understand emotions [68], and body posture can help us with predicting action [141,156]. As demonstrated by a previous study, deep artificial neural networks have been shown to classify sports with greater accuracy by using entire sports videos instead of just a single frame [69]. Conversely, natural vision is rooted in the continuous flow of video rather than discrete image frames. The importance of classifier models trained on videos has significant implications for cognitive neuroscience and is a crucial goal in Artificial Intelligence science [150].

---

[*] Corresponding Author

Video understanding was initially considered an extension of the previously studied and perfected image understanding models. The first models trained by the videos performed the spatial feature extraction in different video frames [4] or used intermediary media such as Optical flow [61] to map the motion and temporal features [45,120,155]. Later models, however, will treat the video as a whole and will allow the model to extract spatiotemporal features directly from it [67,138]. Inspired by the natural vision, multi-stream networks were proposed to implement this spatiotemporal feature extraction on parallel but separate streams [43,122,143]. In these models, one stream would focus more on spatial information extraction while the other would specialize in extracting temporal information. The latest and state-of-the-art models in video understanding are transformer models [127,140] that are experts in data fusion and spatiotemporal feature extraction in their self-attended feed-forward structure.

What kind of information can we take from videos, and how should we use them to result in more accurate learning and understanding of this data? How do we use videos efficiently to train our models? What challenges do we face when we deal with videos instead of images? How can we deal with these challenges without losing too much information? These questions were the main concerns of many researchers in recent years regarding video understanding, and they put comprehensive effort into analyzing and proposing different methods and tools for video classification tasks. In this survey, we will first review the basis of video understanding tasks and spatiotemporal feature detection. This paper provides a holistic overview of the video understanding models. Focusing on the spatiotemporal feature detection techniques, we will discuss a few structural designs and implementations in the video understanding models. We will also discuss the most important benchmarks for video understanding and action recognition and report the results for some of the discussed models for comparison. Finally, this survey details some challenges in spatiotemporal video understanding deep neural network models and some proposed solutions with an outlook to future works.

## 2 VIDEO UNDERSTANDING

We must detect objects and their associated actions to understand and describe the surrounding environment, which necessitates the recognition of relevant nouns and verbs, along with their interconnections, to facilitate video understanding, prediction, or narration. Comprehending videos involves categorizing and grasping the entities, events, and actions depicted within them. We can break it down into two well-established and extensively studied domains: object recognition and action recognition.

### 2.1 Object Recognition

Object recognition and classification is a well-known and well-discussed subject [96]. Our goal for this task is to identify and categorize objects in the given image or video [92]. Many algorithms and neural network architectures were suggested and used for this task. From them, we can mention algorithms that will classify objects based on their appearances (with edge and corner detection or gradient-based methods) [87,116], genetic algorithms [84,95], and algorithms that work based on feature detection [8,88,93]. Features serve as crucial data that provide us with essential insights into the most significant aspects of the acquired data.

Spatiotemporal feature extraction and analysis draw inspiration from natural vision. Some recent tools, such as deep neural networks, can resemble human vision in the task and even outperform humans [72]. Natural vision depends on feature extraction, changing the representation in various steps, and creating a complete understanding of our surroundings and the entities around us.

Artificial models can rival natural vision in terms of accuracy. However, computer models can not effectively handle image variances compared to natural vision. Variations include changes in lighting, rotation, perspective, obstacles blocking the view, and depth. Since earlier methods, computer vision strived to create change invariant models based on the environment [92]. Even with that in mind, some variations remain beyond the scope of image-based object classification methods. It's important to note that some image-based classification methods may not be able to capture changes that occur to an object over time. This failure is because they have no access to temporal data that could help them better understand how an object changes and adapts over time.

### 2.2 Action Recognition

Identifying the verbs that describe the changes or lack of changes in the scene is action recognition [115]. Action detection can extend to static images, potentially yielding improved results when motion does not significantly contribute additional information. In such cases, a single frame can effectively capture the essence of the action (Figure 1).

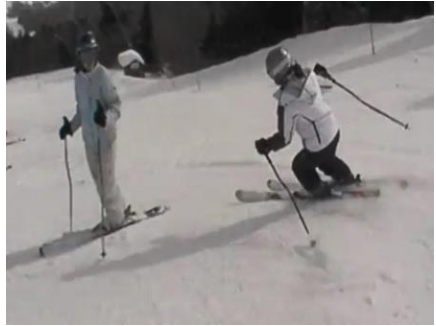

Figure 1: An action might be deductible from a single frame. If the video frame is selected carefully, the model might be able to predict the entire action without having any motion information.

However, in more complicated action recognition and classification tasks, using video for feature extraction and training the network has demonstrated superior outcomes [45]. In these intricate tasks, actions are imperceivable and not categorizable unless observed across multiple frames.

Feature detection is the first step in recognition tasks. We can classify and categorize different entities and actions in the environment with better features that characterize the changes and can describe them. As we need to include temporal information for recognition, spatial feature extraction will not be enough. Because of this, many introduced methods extract spatiotemporal features that can model time and space at the same time for the recognition tasks.

## 3 SPATIOTEMPORAL FEATURES

The first question centers on identifying the categories of data observable within a provided video and what specific elements should receive increased focus. Historically, vision models have predominantly aimed to extract spatial features for image classification. Traditional feature detection methods such as Canny [15], Harris [56], or other conventional algorithms designed for detecting edges, corners, and blobs are usable for spatial feature extraction. Theoretically, a significant portion of the spatial information within images resides within these identified features. Consequently, with these extracted features, we can recognize and classify them with remarkable efficiency.

With the emergence of novel machine learning methods, computers took on the task of feature extraction. These systems would train on data, enabling them to learn the most pertinent details about images relevant for classifying them. This approach allows the model to become specialized and pick up on nuances we might not instinctively identify as significant. The model will enhance its proficiency in extracting more relevant features [128].

This methodology revolutionized object detection and object tracking, which previously relied on manual feature detection. Soon after, with the creation of neural network models and larger datasets such as Image-net [29], analyzing and comparing the different methods for object detection and image classification became possible. Deep neural networks started to show their true potential in feature extraction and classification tasks [57,74].

However, videos introduce an additional dimension: time. The existence of this new dimension for feature extraction compels us to adopt one of these three different methodologies for feature extraction (Figure 2):

**Spatial feature extraction:** to extract the spatial features of the variant frames through time and utilize them collectively as input
**Spatial and temporal features extraction:** to extract the spatial features from different frames and the temporal features (reflecting changes of the spatial features over time) separately, incorporating them into two distinct streams, and utilizing both as the input for the classifier.
**Spatiotemporal feature extraction:** to extract features encompassing all dimensions from the video. These novel feature, which encapsulate information for both spatial and temporal dimensions, are referred to as spatiotemporal features.

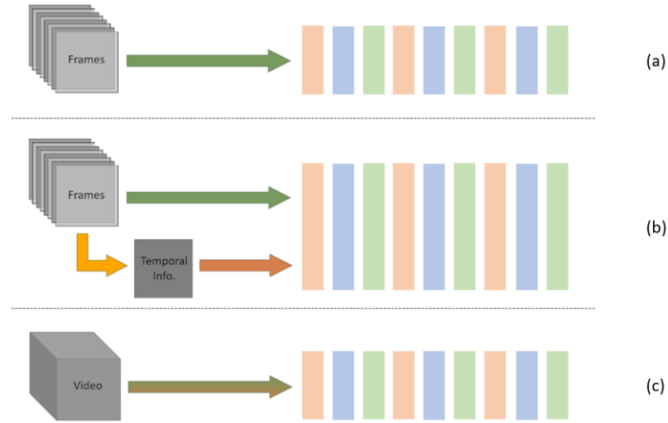

Figure 2: The different networks require different inputs for their classifications. Type a networks take the image frames from the video. Type b networks create a temporal representation (such as Optical Flow [61]) from the video and input both the video frames and this intermediary representation. Type c networks take the entire video as a single integrated input.

## 3.1 Spatial Feature Extraction

In the initial approach, we identify spatial features within distinct frames and employ these spatial features from all frames as input. To achieve this, we choose a single frame as the primary input or combine all frames into a unified image to act as the input. With this kind of feature extraction, we lose the "time sense" of the data. As a result, much information deduced from the temporal channel will be lost [4].

## 3.2 Spatial and Temporal Features Extraction

Creating separate intermediary media that can represent the motion in different sections of videos can assist the temporal feature extraction. Optical Flow [61] is an example of those intermediary media created and used frequently in the introduced methods for representing temporal features. Leveraging tools like Optical flow [61], Object flow, or any other intermediary media that can show the motion to process on a separate and parallel stream [45] alongside the network to extract spatial features has consistently yielded noteworthy performance enhancements across various architectures. This approach represents a straightforward yet impactful means of incorporating pertinent motion data into specialized tasks [60]. Rather than undertaking the non-learnable task of generating these representations of the media, in later models, a distinct and trainable artificial neural network stream is employed in parallel to the spatial stream, with the specific purpose of temporal feature extraction. We will review these multi-stream models in subsequent sections.

There is a potential that the linkage between spatial and temporal data gets disrupted when segregating the temporal and spatial features into two separate streams. For example, while we can identify objects and track their movements, the concept of motion derived from temporal and the recognition of entities through spatial features might remain detached and unrelated.

Earlier models had trouble with the fusion of the spatial and temporal details as the data each stream is modeling is complex. Convolutional neural networks, however, show good potential in linking and fusion of separated data in multi-stream models. Deep neural models have enough parameters to model such fusions and interpret spatiotemporal details.

In deep neural networks, comprehending and extracting deeper insights from data involves a series of transformations towards higher cognitive levels within each layer, altering the data representation through multiple stages [83]. These higher-level representations can assist us with extracting semantics from the input multimedia and subsequently classifying them. The feature extraction primarily revolves around spatial aspects when dealing with a static image input. The neural network enhances these spatial features and generates a new representation of the data in each layer. Data representation undergoes multiple changes until the model achieves its goal. The network then generates a semantic prediction, which we strive to minimize the error during the training phase.

When analyzing videos, depending on the feature extraction methods, we possess distinct spatial features from various frames alongside temporal features that capture changes across the duration of the entire video. Extracted features should also be augmented with each other on both spatial and temporal perspectives in multiple steps to create a new set of features in each step that represents a higher level of semantical information about the input. However, as the depth of our model increases, there's a risk of losing the connection between spatial and temporal information within the network's layers. This concern becomes more pronounced when these two types of information are not consistently intertwined and synchronized over time. Temporal and spatial information processed separately must combine in different parts of the architecture to build spatiotemporal understanding. In preceding models, this fusion would typically occur toward the final stages of the respective streams.

By using multiplication residual blocks and gating, we can constantly send feedback from one of the streams to the other and help the network to go deeper [44]. Adding small feedback channels between the multi-stream will allow them to integrate and fuse interpretations of data earlier and in earlier stages of the model.

### 3.3 Spatiotemporal Feature Extraction

The final method is extracting the spatiotemporal features directly. The spatiotemporal features refer to the information (variance) we obtain from both the location and time, which provide the most relevant data for video understanding. In these networks, we see video as raw data on three dimensions of (x, y, t). We employ 3-D convolution and other associated processes to extract features and execute classification tasks [138]. From this perspective, extracting spatiotemporal features can be likened to extracting spatial features in a higher dimension. From this standpoint, time is akin to an additional input dimension. Nonetheless, this perspective doesn't universally align with feature extraction tasks, as time significantly diverges from spatial dimensions. What we seek from the time dimension extends beyond merely capturing the presence of objects over time; it encompasses motion data and the dynamics of change between distinct frames.

Models emphasizing spatiotemporal feature extraction demand additional time and memory resources for training and recall processes, often grappling with overparameterization, particularly when handling extended video inputs. The removal of longer-term relationships between frames can lead to the loss of critical temporal information. Segmenting videos and generating shorter snippets can significantly compromise the accuracy of this task, as illustrated in (Figure 3) [138].

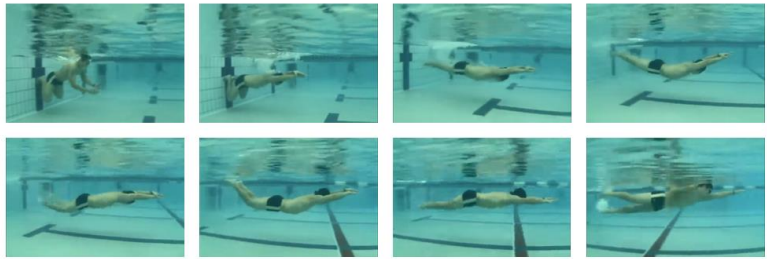

Figure 3: Studying nearby frames is not sufficient to distinguish certain actions, such as swimming movements. As illustrated above, the first four frames can't classify this swimming move. Long-term dependencies are required to detect actions that happen through long time intervals, and models must study more frames as input.

For the extraction of spatiotemporal features, 3-D convolutional layers can be employed [67]. These 3-D convolutions are adept at detecting spatiotemporal features using a dynamic kernel that traverses the current frame and the adjacent frames, thereby shaping the input for the subsequent layer, as illustrated in (Figure 4).

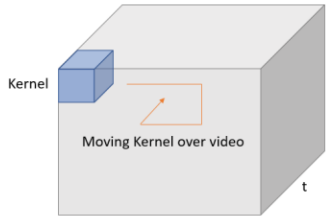

Figure 4: With a 3-D convolution, we move a Kernel over the video to extract spatiotemporal features. This kernel will sample both spatial and temporal relationships.

In 3-D convolutional networks, it has been observed that segregating spatial and temporal fusion leads to a noteworthy enhancement in network accuracy. One approach to achieve this is by decomposing a 3-D spatiotemporal convolutional layer into a 2-D spatial convolution followed by a 1-D temporal convolution, as depicted in (Figure 5) [107,137]. This strategy augments the nonlinearity within the network, thanks to the additional ReLU activation function situated between the 2-D and 1-D convolutions within each block. Consequently, the network, functioning as a function approximator, gains the capacity to model more intricate scenarios [20]. Additionally, this separation simplifies the optimization of weights and fusion processes [123].

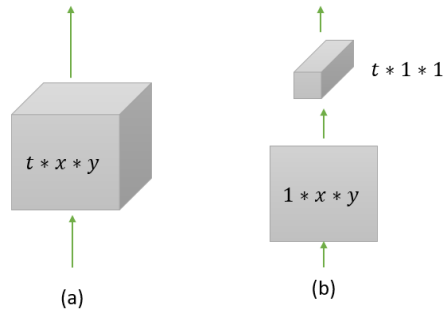

Figure 5. 3-D convolution (a) and (2+1)-D convolution (b) model. Spatial dimensions, including width and height, denoted as x and y, and the temporal extent represented by t, characterize our filter. Figure adapted from [137]

[135] introduced C3D as a generic video descriptor founded on 3-D convolutional networks. Through empirical evidence, the authors demonstrated that employing homogeneous 3x3x3 filters consistently outperforms varying the depth of temporal filters. C3D's descriptor is derived by averaging the outputs from the initial fully connected layer within the network. Furthermore, ongoing research has contributed to the refinement of spatiotemporal convolutional layer structures, yielding significant developments such as the Channel-Separated Convolutional Network [136] and Mvfnet [153].

## 4 UNDERLYING PRINCIPLES OF VIDEO UNDERSTANDING NETWORKS

To systematically categorize the diverse range of models developed for video understanding, it is imperative to delve into critical advancements and discussions that underpin the creation of these models.

### 4.1 Preprocessing Videos

Preprocessing plays a pivotal role in effective learning from videos, and its significance can be attributed to several key reasons:

1. **Selective Information Relevance:** Videos often contain extensive spatiotemporal information, much of which is only partially relevant to the learning task at hand. A substantial portion of this information can be extraneous and is a distraction.

2. **Mitigating High-Dimensionality:** Video data typically comes in a high-dimensional format, which can introduce various architectural challenges in neural networks. Dealing with such high-dimensional data can make it harder for models to generalize learning effectively.
3. **Handling Noise and Distortion:** Videos are susceptible to various sources of noise and distortion. Spatial noise may involve adding random noise to images, applying transformations, introducing obstructing objects, altering the color and shape of objects, or other environmental changes. Temporal noise can arise from edited videos, disrupting the semantic flow of frames (e.g., injecting unrelated frames). Additionally, the use of multiple cameras for recording can introduce temporal variance. These issues collectively constitute the input variance problem, and deep neural networks are inherently sensitive to such variations.

Preprocessing steps are essential to guarantee that the data is in an appropriate form for analysis and learning. These steps are vital for avoiding the aforementioned challenges and ensuring that the neural network can effectively learn and generalize from the video data.

There are numerous types of preprocessing techniques designed to address these concerns and prepare video data for further analysis and modeling. Certain techniques, such as the **Fovea Stream**, were introduced to concentrate on the most pertinent parts of videos, which contain more relevant data for our processing tasks. The core concept is to direct attention to the inherently significant regions within frames where most of the changes occur. This concept is realized by focusing on the central portion of all frames while discarding the peripheral edges due to camera bias. Camera bias arises from the tendency of cameras to prioritize the central area when capturing video footage, as this region typically contains the most detailed information within the frame.

In this technique, to ensure that context information from the entire image is not lost, a lower-resolution base video also exists alongside the Fovea Stream, referred to as the **Context Stream** (Figure 6). This approach effectively reduces the overall input size by half by reducing the context stream's resolution to one-fourth of the original value and downsizing the frame dimensions by half in each dimension for the Fovea stream [69].

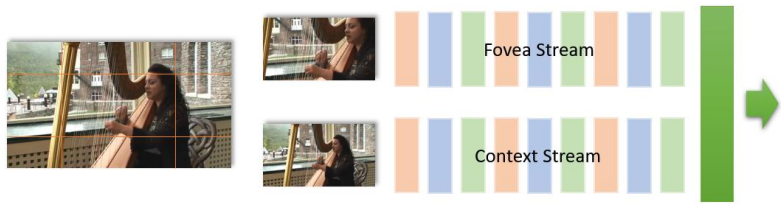

Figure 6: Fovea and Context Streams are parallel similar streams in vision networks. The Fovea stream has the center of the image as input when the context stream reduces the resolution. This model emphasizes the center of the image because of the camera bias. Figure adapted from [69]

Numerous preprocessing techniques have centered around the reduction of input data size by altering data dimensions. Variance, encompassing both spatial and temporal aspects, serves as a potent representation of data changes and reveals information that aids in distinguishing entities and actions. Consequently, variance plays a pivotal role in determining the dimensions for newly represented data.

Dimension reduction methods, such as Principal Component Analysis (PCA), have gained popularity as preprocessing steps for neural networks designed to operate with video data. In some video datasets, like Youtube-8M, videos have already undergone PCA and preprocessing [102], further emphasizing the utility of dimension reduction techniques in preparing video data for neural network-based analysis.

Another critical objective in preprocessing is noise removal, aimed at addressing the input variance problem. In this context, noise refers to any undesired disturbance that impacts the quality of spatiotemporal features. As neural network architectures train with a finite set of supervised data, they can become sensitive to these unwanted features, potentially diminishing the network's ability to generalize effectively.

These disturbances typically manifest as spatial noise and can be managed using conventional noise detection and removal methods, much like those applied to single images. However, videos introduce the additional challenge of temporal noise. Temporal noise occurs primarily when the semantic flow of the video undergoes shifts, often due to changes in the camera perspective or manipulations of the temporal sequencing of frames. Addressing both spatial and temporal noise is crucial for enhancing the robustness and generalization capabilities of neural networks when processing video data.

### 4.2 Temporal Frame Fusion

In a broader context, spatiotemporal convolutional neural networks exhibit a relatively rigid temporal structure, primarily because these networks require a predefined number of frames as input. To imbue the network with temporal awareness, it becomes necessary to either fuse information from multiple video frames or extract features from different frames. The specific approach to this fusion can vary, occurring either before or after the extraction of spatiotemporal features within different models, contingent on their defined structural architecture.

The central question revolves around the selection of frames for fusion. In earlier proposals, the fusion process involved combining adjacent video frames to generate higher-dimensional data that encompassed both short-term temporal information and the spatial data of the video frames. However, it's important to note that the fusion of neighboring frames is not the exclusive or necessarily the optimal approach. [69] introduced and explored several primary types of frame fusion methods, as illustrated in (Figure 7).

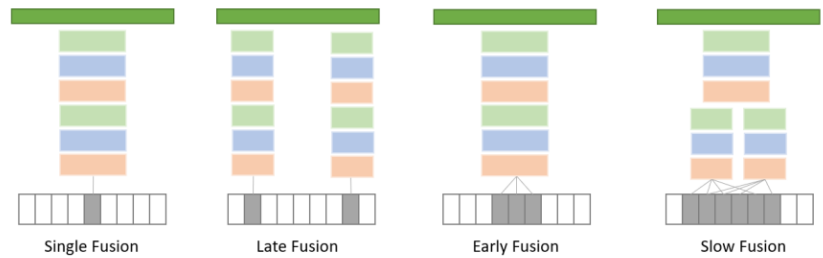

Figure 7: Different types of fusion, from left to right: Single frame with no fusion, Late fusion of related video frames, early fusion, and slow fusion. Single fusion process a single frame. Late fusion combines the information from distant frames. Early fusion combines the info in close-by frames. Slow fusion uses a hierarchical approach for combining the frame data. Red, blue, and green boxes indicate convolution, normalization, and pooling layers in order. Figure adapted from [69]

1. **Early fusion** is akin to the fusion that occurs in traditional 3-D convolution-based methods [67]. In early fusion, a new representation is generated that places a stronger emphasis on preserving the spatial data without significant alteration. This approach is effective for detecting short-term temporal features but may be relatively less proficient at recognizing and utilizing long-term temporal relationships between different frames. Early fused networks tend to struggle to extract long-term features or leverage long-term temporal patterns to comprehend the content of the video. Their primary strength lies in capturing short-term dynamics and spatial information within the video frames.
2. **Late fusion** involves the fusion of frame-wise features occurring at a later layer in the network. This approach enables the network to effectively handle long-range relationships between spatial features. However, one notable challenge with late fusion networks is that there is no guarantee that long-term relationships between frames always exist in videos, particularly in web-based videos. In online videos, it's possible to switch scenes abruptly and present unrelated and distinct video frames. Additionally, temporal shifts can occur, further complicating the establishment of semantic relationships between non-neighboring frames. As a result, late fusion networks may encounter difficulties handling such scenarios where long-term relationships between frames are not readily discernible.
3. **Slow fusion** involves the sequential processing of several consecutive sections within videos using the same network structure. This fusion approach repeats hierarchically in a pyramid-like fashion, as depicted in (Figure 7), to develop a comprehensive understanding of the entire temporal domain. Notably, slow fusion has demonstrated superior performance compared to other fusion methods, making it an effective choice for capturing and utilizing temporal information in video data. You can find the results of this comparison conducted by [69] in Table 1.

Table 1: Comparison of different fusion methods on 200000 videos from the Sports-1M dataset. Hit@K determines the fraction of test samples containing at least one of the ground truth labels in their top K results. Results from [69]

| Model | Clip Hit@1 | Video Hit@1 | Video hit@5 |
|---|---|---|---|
| Single-Frame | 41.1 | 59.3 | 77.7 |
| Early Fusion | 38.9 | 57.7 | 76.8 |
| Late Fusion | 40.7 | 59.3 | 78.7 |
| Slow Fusion | 41.9 | 60.9 | 80.2 |

From a cognitive neuroscience perspective, slow fusion closely resembles natural vision [12]. In natural vision, recorded and fused frames are processed through various pathways within different brain regions, which leads to the creation of new data representations at each layer. These novel data representations, which vary depending on the brain's recording locations, are associated with different aspects of visual perception. They encompass fundamental properties of images, such as color, angles, directions, and motion, as well as more advanced properties such as object recognition, spatial relationships between objects, and the body's position within the visual space [99].

In the Temporal Relational Network (TRN), a distinctive feature is the selection of various frames to uncover both early and late relationships, with subsequent aggregation of this data in later network layers [172]. Selecting a varying number of frames as input allows the model to switch dynamically between varying types of fusions.

Advancements in fusion methods have led to the development of a pyramidal Gaussian fusion approach. This technique fuses frames from different time intervals and employs a temporal Gaussian attention filter. The key innovation here is the selective emphasis placed on certain video frames while considering the others within each time interval. As a result, the network dynamically switches between early and late fusion approaches based on the context and the specific temporal Gaussian attention filters selected for the task, as illustrated in (Figure 8) [105].

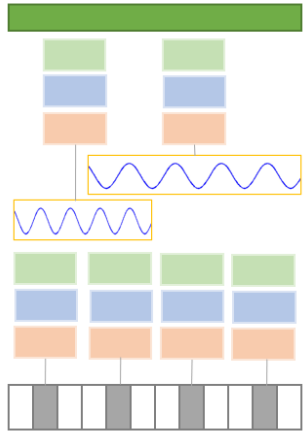

Figure 8. The above figure illustrates pyramidal Fusion with Temporal Gaussian attention filters. In this fusion, multiple learnable Gaussian filters apply to the input frames. Each filter will have a different take on context. The network can benefit from early and late fusions based on how Gaussian filters are set dynamically. Figure adapted from [105]

The hierarchical structure, which combines early and late fusion techniques to utilize short clips' motion for short-term spatiotemporal feature extraction and the entire video's motion for long-term spatiotemporal feature extraction, has found applications in various domains. One notable application is in video question-answering tasks [77]. This hierarchical approach allows the network to leverage short-term and long-term temporal relationships in the video data, enhancing its ability to address complex tasks such as video-based question-answering.

There are other innovative approaches in addition to the previously mentioned temporal fusion methods. Drawing inspiration from natural language processing techniques, the progressive training method (PGT) [100] treats videos as a sequence of fragments that adhere to the Markov property. Leveraging this concept, it propagates information through the temporal dimension in multiple stages. This multi-stage fusion approach enables end-to-end training of deep networks on large video datasets and facilitates the efficient handling of temporal information across extended sequences.

### 4.3 Data aggregation and Pruning

In their research, [163] explored various models for data aggregation, as illustrated in (Figure 9). The fundamental concept behind these models is to transform the data structure and amalgamate information at each layer, ultimately constructing a new feature set that can be effectively utilized in the subsequent layer. This information aggregation process must persist in the later stages of the network. Pooling operations play a crucial role in regulating the creation of the feature set for the next layer, offering control over the network's hierarchical feature extraction. [117] conducted research into the application of Max pooling to generate long-range temporal aggregated data, which can significantly contribute to video understanding. In their study, they devised temporal aggregation blocks (TAB) that rely on attention filters and max pooling techniques. These TABs designed to aggregate information from the recent past and more extended periods will effectively aid in predicting future actions within the video context [117].

Before choosing the pooling method for each section in the model, we must determine whether we should decrease, increase, or maintain the dimensions in that layer. As our network advances, we aim to reduce the dimensionality until a straightforward relationship emerges between the final feature set and our task. In essence, our goal is to streamline the dimensions and link the raw spatiotemporal features to our ground truth labels in later layers.

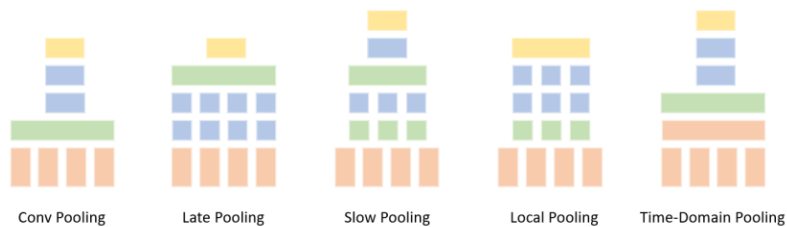

Figure 9: Some of the different pooling methods. Red, Green, Blue, and Yellow blocks show Convolution, Max pooling, fully connected layers, and SoftMax in the same order. Figure adapted from [163]

A commonly employed and well-received approach involves dimension reduction and subsequent expansion, which serves as a data encryption and decryption method. This methodology can be categorized as a pruning technique, where the neural network can remove non-essential data for the classification task. This process aids the network in reconstructing features closely tied to the main context. Autoencoder-based architectures have been widely used in video understanding and classification tasks [101], particularly in applications like anomaly detection and tracking [21,113,170], video compression [47,162], video retrieval [124], and video summarization [167]. In Compressed video action recognition (CoViAR), the model processes the video through spatial and temporal streams before applying any codecs [152].

One of the highly effective data aggregation methods, particularly suited for temporal features, is NetVLAD [2]. The output of the neural network in this approach consists of a set of descriptors, each of which contributes to describing the entire network to some extent. These output descriptors serve as vector inputs for VLAD [3,66]. VLAD, in turn, generates a single vector output that encapsulates the described information from the final layer, including spatiotemporal details from the fused frames. NetVLAD incorporates SoftMax and normalization layers (Figure 10) to address input variance issues. As previously discussed, input variance problems pertain to how slight changes in objects or the video can impact the outcome. NetVLAD is commonly employed in various video classification and comprehension architectures [132]. For instance, ActionVLAD was among the early architectures to utilize the VLAD aggregation method for spatial and temporal streams [46], while the WILLOW network leveraged NetVLAD for data aggregation and emerged as the winner of the inaugural Youtube-8M Large-scale video understanding competition [94].

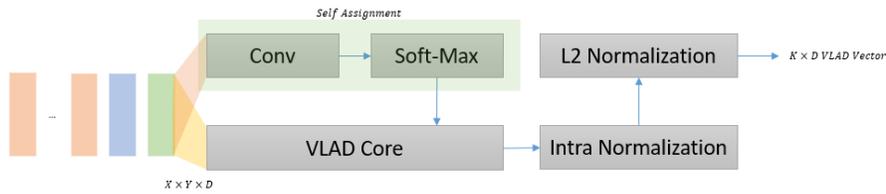

Figure 10. NetVLAD structure. The VLAD core will augment the input vectors and produce a vector representing them all. NetVLAD introduces new blocks for normalization and input invariancy. Figure adapted from [2]

In more recent developments, a technique known as Selective Dependency Aggregation (SDA) [130] has emerged. SDA is compatible with various backbone structures and is designed to identify both long and short-range features within videos. It achieves this by employing multi-directional and multi-scale feature compression and dependency enhancement based on the inherent dependencies present in the video data.

Effective aggregation and parameter reduction can lead to a more compact network that requires fewer Floating Point Operations per Second (FLOPS). An example of the importance of memory efficiency can be seen in the second ECCV workshop on Youtube-8M Large-scale video understanding, hosted by Google AI in 2018 [78]. In this competition, participants were challenged to create models that use less than 1 gigabyte (GB) of memory. The goal was to promote memory-efficient single models rather than relying on large ensemble models. One of the top-performing architectures in this competition was NeXtVLAD [86], which modifies the NetVLAD approach to reduce the model's dimensionality.

Another approach is Teacher-student networks, which select specific frames essential for video understanding tasks. These models effectively reduce the number of FLOPS, computational costs, and overall model size [11]. In the process of extracting spatiotemporal features for video classification, videos are often divided into shorter clips. The network then combines predictions made at the clip level to generate predictions for the entire video. However, this approach can lead to redundancy in the extraction of spatiotemporal features since visually similar clips within the video are processed separately. To address this issue, FASTER (Feature Aggregation for SpatioTemporal Redundancy) is a recurrent model that can process up to 75% of the clips using a much more computationally efficient model without sacrificing accuracy [173].

### 4.4 Attention and Shifting

In various layers of the network, we apply different attention filters to the input data. These attention filters have varying weights and are trainable, meaning the neural network can learn to prioritize specific parts of the previous layer's data, which enables the model to determine which aspects of the previous layer's data are more important and warrant closer attention (Figure 11) [120].

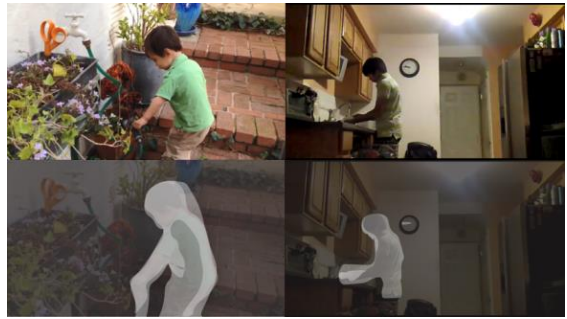

Figure 11. Attention is applied to focus the network on the action. Based on Optical flow [61] or other media that can map the changes, the model can focus on actions such as watering the garden and washing dishes.

Attention-based neural networks have found applications in a wide range of recognition and classification tasks, including sound recognition, natural language processing, and image recognition. These models have demonstrated their effectiveness in improving performance and handling various types of data.

Shifting is a technique introduced to simplify attention complexity. It works similarly to attention but focuses on streamlining the data to emphasize the main context. In this approach, attention weights are transformed into a straightforward shift operation. This means that the maximum weight in an attention filter is rounded up to one, and the rest of the attention weights are set to zero (Figure 12). This reduction in attention complexity also leads to a decrease in the total FLOPS required for the neural network. Shifting is particularly useful when applied to the temporal channel, as seen in models like the TSM model [85]. However, it can also impact spatiotemporal streams in more complex models like RubicksNet (Figure 13) [39].

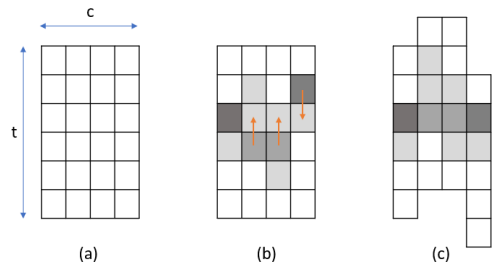

Figure 12. The quantized temporal shift. Images (a) and (b) show the attention window. We simplify the attention on each channel to a single weight of 1 and rest zeros and will treat it like a shift in image c. Figure adapted from [55]

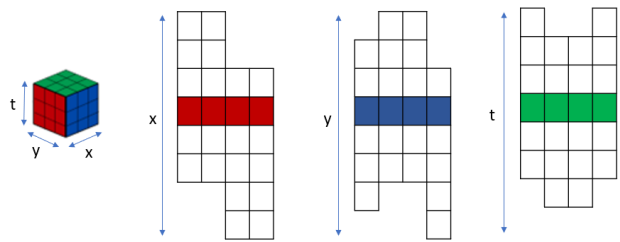

Figure 13. This figure illustrates the spatiotemporal shift in RubicksNet. The shift happens in all dimensions. Figure adapted from [39]

Shifting the time axis can occur in either direction if we have access to future frames. However, if we only have real-time information, we can only shift temporally in one direction, which is backward in time. Spatial shifting, on the other hand, can happen in both directions. To train shift networks effectively, we need an additional loss function. This loss function is essential to penalize excessive shifts that could potentially harm the spatiotemporal integrity of the data. Similar penalties should apply to control excessive changes in other attention-based networks.

Context gating is one of the applications of attention-based or shift-based models. It's used in later layers of the network to enable it to filter out objects or actions that are not relevant to the context (Figure 14). Context gating is valuable for refining the network's output by helping it judge the output's confidence and relevance. It's particularly helpful when combining results from multiple networks to make higher-level deductions [94].

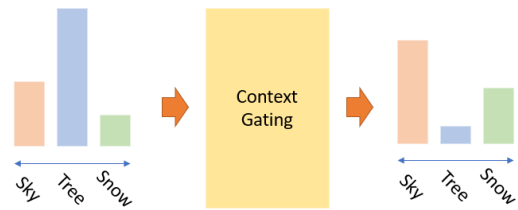

Figure 14. An example of context gating for a video about skiing. Snow and Skiing are more significant for action detection than trees, though the network has more confidence in detecting trees. Figure adapted from [94]

By incorporating context gating and a straightforward follow-up network like a simple RBF kernel, it becomes possible to classify based on the detected features from various channels. The WILLOW network, which emerged as the winner of the 2017 Youtube-8M competition, employed this uncomplicated structure with audio and video data (Figure 15) [94]. Subsequent enhancements to this architecture in 2018 involved distillation methods, average weights, and quantization of weights in different layers [78,86].

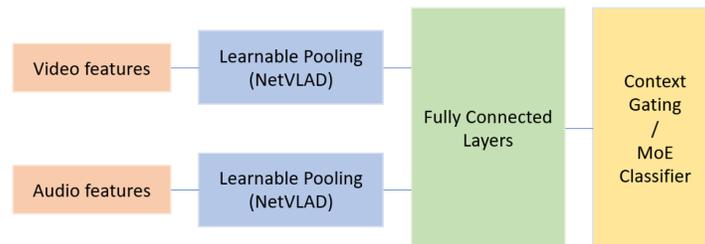

Figure 15. The WILLOW structure includes multiple streams for video and audio feature extractions. The augmented data with fully connected layers pass through context gating before classification. Figure adapted from [94]

## 5  VIDEO UNDERSTANDING MODELS

As in other domains of computer vision, a range of methods have been devised to build models for video comprehension. The next section will review fundamental approaches, considering feature extraction and overall architectural perspectives.

### 5.1  Structural Search Models

One of the initial and rewarding approaches to video comprehension involved extending image understanding models into the temporal dimension. This approach applied structural search techniques, previously employed in image understanding, to convolutional feed-forward architectures for video comprehension. The process began with a simple model, which progressively expanded and was analyzed for performance to determine its optimal state.

In pursuit of improved structures for video understanding tasks, [158] searched for various network structures featuring 3D and 2D convolutional layers to strike a balance between speed and accuracy. The X3D network, employing Neural architecture search and stepwise network expansion techniques, systematically extended a basic 2D image classification network along multiple axes, including spatial, temporal, weight, and height dimensions of different blocks, which resulted in a family of efficient networks tailored for video understanding and classification [42]. Similarly, AssembleNet [112] and AssembleNet++ [111] introduced a generic structural search approach to learn connections among feature representations across input modalities, delivering outstanding performance.

### 5.2  Memory-based and Recursive Models

The results of the proposed algorithm strongly suggest that training should encompass the entire video rather than relying on short clips [163]. While models can detect many actions by analyzing a continuous sequence of frames or by extracting short-term temporal features and spatial features from related video frames, there are instances where understanding actions necessitates linking events across time. Moreover, changes in the scene can disrupt the semantic connection between nearby frames.

There are numerous scenarios where a video loses its short-term semantic coherence. In such cases, a frame may exhibit a weaker semantic correlation with its neighboring video frames. Consequently, observing video frames over an extended duration can offer deeper insights into specific events, rendering the details more significant (Figure 16)[151]. Given these challenges, it is advantageous to retain long-term spatiotemporal features, especially when working with longer videos (rather than short snippets), web videos, or videos depicting tasks with notable long-term dependencies that can enhance action understanding. Late fusion and memory-based models are better suited for capturing long-term spatiotemporal features in such scenarios.

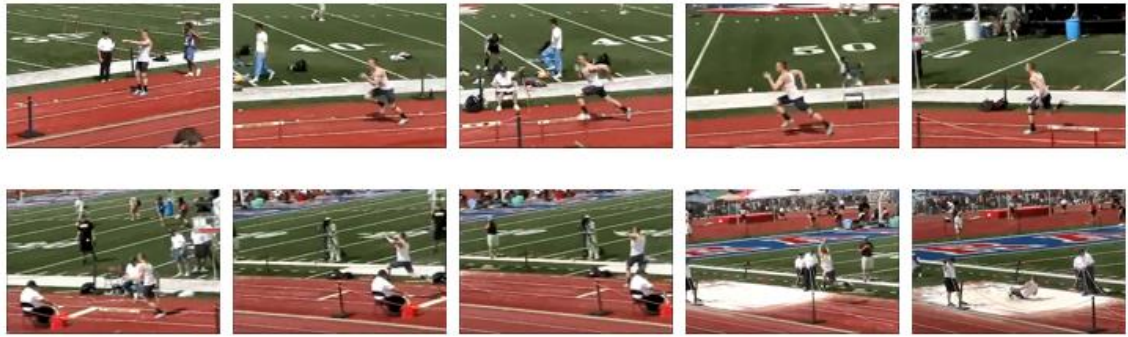

Figure 16. The above image shows a 5-frame window that does not include any clues about the main action that is taking place. With five more frames, we can distinguish between the different sports of running and jumping and their variations.

Memory-based network models, which utilize architectures like LSTM [51,59,129], GRU [24], and other networks [63], effectively leverage temporal information and serve as long-term feature repositories. These structures are highly efficient and valuable when dealing with temporal features or when patterns evolve, as seen in tasks like speech recognition [50]. (Figure 17) is a visual representation of the typical structure of memory-based and multi-layer recurrent networks employed in previous video understanding works [6,33].

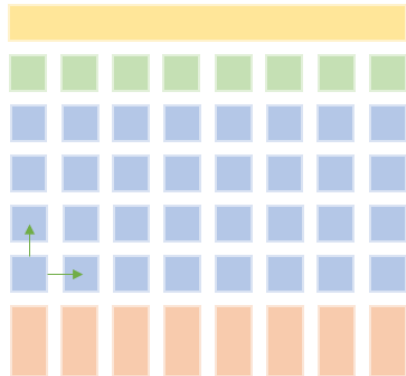

Figure 17. General Recurrent Neural Network Structure for video understanding with memory blocks. Red blocks show convolutional layers that receive input after frame fusion. Also, each blue block shows a memory block like LSTM. Figure adapted from [163]

Another approach to capturing long-term features involves creating a separate feature bank alongside the spatiotemporal network. This temporal feature bank offers a comprehensive, time-wise understanding of the entire video and can influence either the spatiotemporal stream or the classifier. While the feature bank excels at extracting long-term temporal features, it is comparatively less proficient at capturing spatial features. On the other hand, the spatiotemporal stream focuses on nearby frames with a smaller window, making it more adept at handling spatial features and short-term temporal features. (Figure 18) provides a visual representation of this concept [110,151].

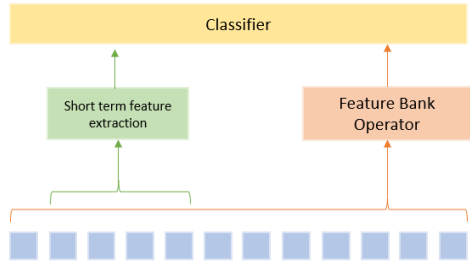

Figure 18. Using feature banks alongside the spatiotemporal stream for better classification. Feature banks will map long-term dependencies. Figure adapted from [151]

Recent developments in the detection of long-term and non-local dependencies have led to the creation of non-local neural networks. These networks assess the relationship between each position in a chosen frame and other parts in different video frames [147]. For instance, CCNet enhances the non-local network's performance by incorporating two crisscross blocks [64]. Meanwhile, GCNet, which employs blocks similar to those found in the Squeeze and Excitation Network [62], offers superior global context modeling capabilities compared to traditional non-local networks [16].

Temporal modeling encompasses both short-range motion detection and long-range aggregation to capture temporal dependencies. In addition to recursive, memory-based, and feature bank-based approaches, there are feedforward models with different in-block temporal fusions designed to handle short-term and long-term temporal dependency extraction. For example, the Temporal Excitation and Aggregation Network (TEA) employs Motion Excitation (ME) blocks for short-range motion modeling, followed by Multiple Temporal Aggregation (MTA) blocks for long-range aggregation [82]. The ME block conducts an early temporal fusion between neighboring frames along the time dimension, while the MTA block performs a late temporal fusion.

### 5.3 Multi-Stream Networks

In our brains, natural vision is processed through two distinct parallel pathways: the dorsal and ventral pathways. These pathways originate in the occipital lobe and progress sequentially toward the parietal and temporal lobes. They are specialized with distinct structures for feature detection, ultimately resulting in a comprehensive vision perception after data augmentation in association areas of the brain. The ventral pathway, which extends towards the temporal lobe, specializes in semantic analysis and object detection. Conversely, the dorsal stream is specialized in motion processing, action recognition, and understanding the spatial relationships between objects and the body and other entities (Figure 19) [48].

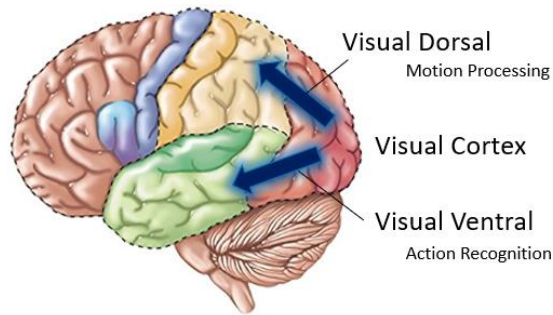

Figure 19. Dorsal and Ventral visual streams are two parallel pathways for natural vision. The visual dorsal is tasked with motion processing and augmentation with sensory perception, while the visual ventral is tasked with object recognition and semantic analysis.

The concept of these two brain streams has inspired neural network models for video understanding, incorporating two parallel streams. In these network architectures, one stream focuses on temporal analysis and feature extraction, while the second stream is dedicated to spatial feature detection. An early example of such a network is the two-stream network [122], which featured two separate visual streams. This model effectively fused both spatial and temporal information, making it a state-of-the-art approach at the time. Drawing inspiration from traditional handcrafted methods that rely on trajectory-based feature extraction, the TDD model [143] is another two-stream network that operates with aggregating trajectory-pooled features.

In subsequent years, newer two-stream networks were introduced, featuring frame fusion and memory-based architectures [45,155]. The stream responsible for spatial analysis can operate on a single frame or a short sequence of frames to generate short-range spatiotemporal features. Conversely, the stream dedicated to temporal analysis can consist of a simple network with inputs such as optical flow [61] or other motion-based images. Alternatively, this stream can employ a 3D convolutional network or another fusion-based architecture designed to detect spatiotemporal features.

Two-stream networks initially faced challenges when dealing with longer videos. To address this limitation, [144] introduced the Temporal Segment Network, which featured a two-stream structure capable of working with long videos through segmentation. Similarly, the deep Temporal Linear Encoding (TLE) model adopted a segmentation approach, breaking the video into different segments and performing two-stream feature extraction for each segment. The results were aggregated and encoded, followed by fusion [32]. One significant issue with the original two-stream networks was their reliance on intermediary motion representations like optical flow [61], which incurred a high computational cost even after training. Subsequently, multi-stream networks were proposed that eliminated the need for these intermediary representations [131,175].

The SlowFast network [43], draws inspiration from the natural vision pathways, utilizing two streams to extract spatiotemporal features. One of these streams, referred to as the Fast stream, processes high frame rate input with extensive temporal coverage. It is primarily focused on motion and time-based analysis. In contrast, the other stream, known as the Slow stream, handles low frame rate input with reduced temporal dominance. The Slow stream specializes in capturing categorical semantics and places more emphasis on spatial information. Importantly, the Fast stream provides continuous feedback to the Slow stream at different layers. This feedback mechanism enables the Slow stream to leverage motion information effectively (Figure 20).

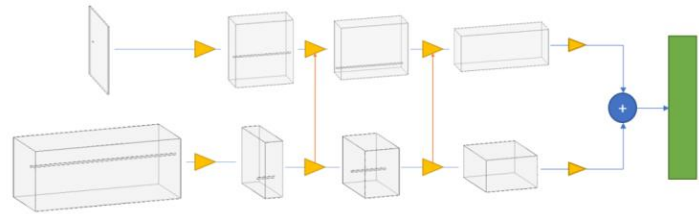

Figure 20. The Slowfast structure with slow and fast streams. The red arrow shows feedback from slow to fast stream. The blue circle shows context gating and fusion of the streams. The green layer represents an RBF classifier. Figure adapted from [43]

In the process of making the final prediction, the output results from multiple streams are aggregated before the final classification. One approach to achieve this is the Mixture of Experts classifier, which employs an attention-based method. In this approach, specialized expert classifiers are trained for each stream, and gating weights are assigned to each classifier. These weights determine the significance of the results produced by each expert classifier based on the content. Following training, the network focuses on the results from various channels to generate a more accurate output, relying on the most significant findings for categorizing the input (Figure 21) [7]. Expandable structures designed to work with a vast collection of long videos can be trained using a combination of expert classifiers. This scalability enables the aggregation of data from a wide range of sources [65].

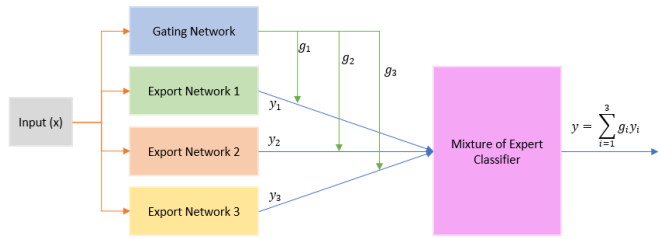

Figure 21. A Mixture of expert classifiers with an attention-based gating network will combine the results from each export network. The mixture of expert classifiers uses a gating stream to determine the importance of each expert stream based on input and will have this gating network trained alongside the expert streams. Figure adapted from [7]

### 5.4 Transformer Networks

Transformer models have recently exhibited outstanding performance in numerous natural language processing tasks. Prominent examples of these models include BERT (Bidirectional Encoder Representations from Transformers) [30], GPT (Generative Pre-trained Transformer) [108], RoBERTa (Robustly Optimized BERT Pre-training) [90], and T5 (Text-to-Text Transfer Transformer) [109].

Transformer models are a class of neural networks characterized by their feedforward encoder-decoder architecture, relying on a mechanism called self-attention. Unlike recurrent networks that process elements sequentially, transformers can simultaneously attend to entire sequences, enabling them to capture long-term dependencies in data. However, this capacity comes at the cost of a substantial number of parameters. For instance, GPT-3 [13] boasts an impressive 175 billion parameters, while the switch transformer model [41] takes it even further, scaling up to a staggering 1.6 trillion parameters.

Transformer models employ a distinctive architecture founded on attention mechanisms. Self-attention, a core component, assesses the significance of each element in a sequence concerning all others. To illustrate, when processing a set of words, self-attention determines the likelihood of a word being associated with other words. In a self-attention block, three adaptive attention vectors rescale input values and generate a consolidated representation for the subsequent layer (Figure 22). Multiple self-attention blocks are concatenated to form each multi-head attention block (Figure 23). Multi-head attention is designed for efficient parallelization, enabling the network to perform context gating in each layer.

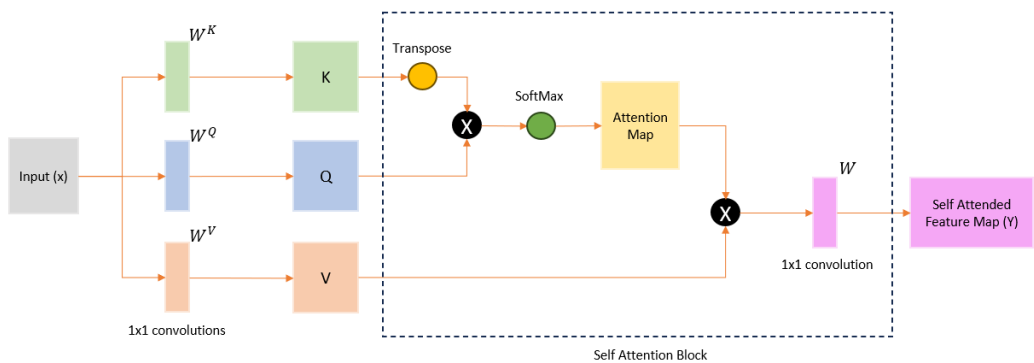

Figure 22. Self-Attention Block. The triplet (Key, Query, Value) applies to input and will rewrite the values. Finally, Output projection (W) applies to the data to transform it. Figure adapted from [71]

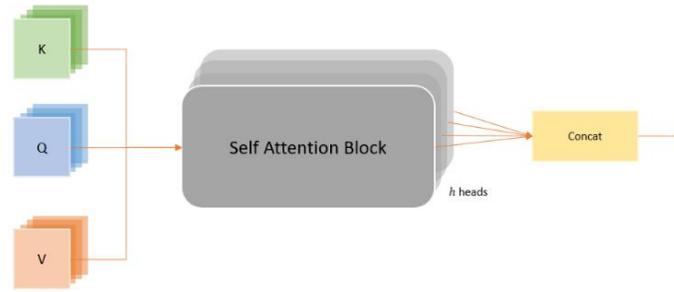

Figure 23. Parallel self-attention blocks create a multi-head attention block. The results concat and are sent to the next layer. Figure adapted from [71]

Research has established that multi-head self-attention when equipped with an adequate number of parameters, offers greater versatility compared to other convolutional operations. Empirical findings from these studies [27] indicate that multi-head self-attention blocks excel in capturing both global and local features within sequences. Furthermore, they suggest that these blocks can enhance expressiveness by adaptively learning kernel weights, akin to deformable convolutions [28].

Transformer networks leverage pre-trained modules to acquire an initial model based on patterns. In the realm of natural language processing, these patterns constitute the language model, which encompasses the probabilities and statistics governing the sequential arrangement of words and tokens to construct sentences. This model offers a structural depiction of how acquired patterns can be generated and replicated. Following this initial pretraining, the networks undergo a second training phase, tailored to their intended task and design. This subsequent training enables the model to specialize in its designated task. Consequently, these networks can employ pre-trained models for innovative and progressive tasks, allowing them to generate fresh patterns.

The initial transformer models were auto-encoders (Figure 24). In these auto-encoder models, input tokens constituting a pattern were processed. They encoded this input data and utilized both the encoded data and previous outputs (which were outputs shifted in time) to generate the subsequent output within a feed-forward network. This process gave rise to a semi-recursive, yet fundamentally feed-forward, encoder-decoder network. Consequently, this model took in a pattern as input and generated a new pattern as output, relying on both input and output data.

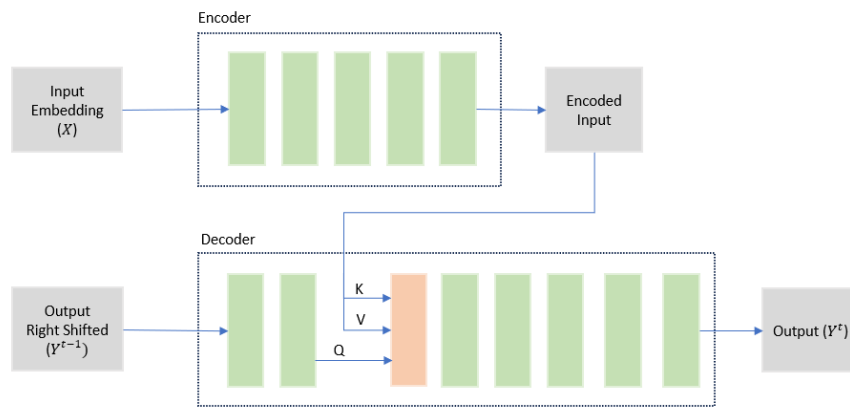

Figure 24. An Encoder-Decoder Transformer model with multi-head self-attention blocks. Figure adapted from [71]

As previously mentioned, the early transformer models employed shifted previous outputs in combination with an encoder to generate new outputs. Based on this structure, these models were unidirectional. In contrast, later transformer models consisted solely of the encoder component and operated in a bidirectional manner, no longer depending on previously generated outputs. BERT (Bidirectional Encoder Representations from Transformers) [30] was a prominent example of a bidirectional encoder-based transformer model that demonstrated exceptional performance across various natural language processing tasks.

Transformer model networks that were originally designed for sequence-to-sequence tasks have found applicability in various computer vision tasks, including video understanding and action recognition. Beyond these applications, transformer models have been successfully employed in a wide range of computer vision tasks such as object detection [17,23,40,149,174], image segmentation [126,157,171], image and scene generation [36,148], image processing [22], super-resolution and scaling [161], colorization [76], clustering [79,80], and 3D analysis [53,169]. These versatile models have showcased their effectiveness across diverse visual processing tasks.

Transformer models are known for their structured and highly parametrized feed-forward architecture, which is typically designed to handle only short video sequences, usually just a few seconds in length. However, as highlighted in the previous sections of this literature review, long-term dependencies play a crucial role in video understanding, particularly for action detection. To tackle this challenge, researchers have explored various approaches to adapt transformer models for processing longer videos in video understanding tasks.

VideoBERT, as presented in the study by [127], introduced a novel network architecture that builds upon the foundations of BERT. It is designed for joint video and language modeling, utilizing masked visual and linguistic tokens. In this model, visual features are extracted from pre-trained CNN video classification models and combined with language tokens as input for BERT, allowing the model to learn the joint distribution of these modalities. Additionally, [89] proposed a BERT-based model for concept localization in videos, which was entered into the third YouTube-8M competition for large-scale video understanding. These efforts highlight the versatility of transformer-based models, such as BERT, in handling complex tasks in video understanding and localization.

The early efforts in action recognition using transformer networks led to the development of models like VTN [98]. VTN employs a two-stage process where it first extracts features using a 2D CNN and subsequently utilizes a transformer encoder network to learn temporal relationships. However, as the field progressed, researchers began exploring end-to-end transformer models for action recognition, incorporating transformer-based feature extraction directly into the process.

Expanding upon the success of the Vision Transformer (ViT) [35] in image understanding, Multiscale Vision Transformers (MViT) [38] introduce a feature hierarchy or fusion approach in videos. This is achieved by progressively altering the spatiotemporal resolution, using multi-head pooling attention. Such alterations in resolution enable the network to effectively model both early and late temporal fusions. TimeSFormer [10], another model based on ViT [35] for video understanding, treats videos as a series of patches extracted from frames. It incorporates spatial and temporal attention divisions in each block. Another notable transformer model in this domain is the Video Vision Transformer (ViViT) [5], also based on the ViT transformer architecture. It's worth noting that ViT-based transformer models often require pre-training with image data to effectively train with videos.

Several state-of-the-art modules, such as TQN [165], LSTR [160], and STAM [119], are also rooted in transformer models. These end-to-end multi-stage transformer models are capable of encoding a sequence of frames and aggregating their results with future sequences, enabling them to handle longer videos despite the rigid feedforward structure of transformer networks. [166] have even introduced a transformer model that relies on token shifting for video understanding.

## 6 EVALUATION AND BENCHMARKING

In the previous section, we explored various structures and their contributions to video understanding tasks. Now, let's delve into benchmark datasets commonly used for video understanding and classification tasks. We will also compare the performance of different networks based on their reported results on these benchmark datasets.

The following table (Table 2) lists key and major datasets for video understanding tasks and classification. As it's observable, the latest datasets had a huge increase in the number of classes and videos (from 7k videos and 51 classes in HDMB51 [75] to over 6M videos and 3862 classes in Youtube-8M [1]). Also, the rate of new dataset production increased in the latest years, at the same time when there was a rapid growth in the number of proposed deep neural networks for these tasks.

Temporal segmentation that is available in some datasets, for instance, YouTube-8m [1], HACS [168], or FineGym [118], provides some information that helps with content/action localization tasks. Spatiotemporal segmentation, like the one provided in the AVA dataset, provides information about where and when in the video the action exists.

Table 2: Some of the key datasets for video understanding and action recognition tasks

| Dataset | Year | Source | # Videos | # Classes | Ave. Duration | Sample classes |
|---|---|---|---|---|---|---|
| HMDB51 [75] | 2011 | YouTube | 7 K | 51 | [2, 3] s | Brush-hair, Kick, Kiss |
| UCF101 [125] | 2012 | YouTube | 13 K | 101 | [2, 5] s | Diving, Skiing, Apple Eye Makeup |
| Sports1M [69] | 2014 | YouTube | 1 M | 487 | 5.5 s | Cricket, Disk Golf, Gliding |
| SVW [114] | 2015 | Smartphones | 4.1 K | 74 | 11.6 s | Archery, Golf, Running |
| ActivityNet [14] | 2015 | Web Videos | 28 K | 203 | [5, 10] m | Gardening, Paying Delivered Pizza, Shoveling Snow |
| YouTube8M [1] | 2016 | YouTube | 8 M | 4800 | 229.6 s | Performance Art, Driving, Aircraft (+ Temporal Segmentation) |
| Charades [121] | 2016 | Crowdsourced | 9.8 K | 203 | 30.1 s | Open Fridge, Hold Shoes, Read Book |
| Sth-Sth V1 [49] | 2017 | Crowdsourced | 100 K | 174 | [2, 6] s | Moving [Something] from left to right, holding [Something], [Something] being deflected from [Something] |
| Sth-Sth V2 [49] | 2017 | Crowdsourced | 220.8 K | 174 | [2, 6] s | Holding [Something], Approaching [Something] with your camera, Closing [Something] |
| AVA [52] | 2017 | YouTube | 351 K | 80 | 15 m | Answer phone, Hug, Drive (+ Spatiotemporal Segmentation) |
| Kinetics400 [70] | 2017 | YouTube | 306 K | 400 | 10 s | Shaking hands, Salsa Dancing, Dribbling Basketball |
| Kinetics600 [18] | 2018 | YouTube | 495.5 K | 600 | 10 s | Vacuuming Floor, Watching TV, Karaoke |
| MiT [97] | 2018 | Web Videos | 1 M | 339 | 3 s | Dancing, Sneezing, Burying (Humans, Animals and objects as agents) |
| Kinetics700 [19] | 2019 | YouTube | 650.3 K | 700 | 10 s | Moving Baby, Peeling Banana, Arresting |
| HACS Clips [168] | 2019 | Web Videos | 1.5 M | 200 | 2 s | Same classes as ActivityNet [14] (+ Temporal Segmentation) |
| AVA-Kinetics [81] | 2020 | Kinetics700 [19] | 230 K | 80 | 10 s | Same classes as AVA [52] (+ Spatiotemporal Segmentation) |
| HVU [31] | 2020 | YouTube8m [1], Kinetics600 [18], HACS [168] | 577 K | 4378 | 10 s | Butterfly, Wedding reception, Long Distance Running |
| AViD [106] | 2020 | Internet | 450 K | 887 | [3, 15] s | Baking cookies, using a paint roller, Chopping wood |
| FineGym [118] | 2020 | Gymnastics Competition Videos | 303 | 530 | [8, 55] s | Gymnastics: Balance Beam: Flight Handspring: Handspring Forward with leg change (+ Temporal Segmentation) |

We compare some of the reported performances for some of these major networks for some of the most used benchmark datasets for video understanding and action recognition (Table 3).

Table 3. Comparison of main introduced structures for video understanding. Mean Average Precision (mAP) for Charades and Hit@1 accuracy for other datasets are used for the reported results. In all cases, the source literature provided the results.

| Method | Year | Pretrain | Backbone | HMDB51 [75] | UCF101 [125] | Charades mAP [121] | Sth-Sth V1 [49] | Kinetics400 [70] |
|---|---|---|---|---|---|---|---|---|
| DeepVideo [69] | 2014 | ImageNet [29] | AlexNet | - | 65.4 | - | - | - |
| Two-Stream [122] | 2015 | ImageNet [29] | CNN-M | 59.4 | 88 | 22.4 | - | - |
| TDD [143] | 2015 | ImageNet [29] | CNN-M | 63.2 | 90.3 | - | - | - |
| C3D [135] | 2015 | Sports1M [69] | VGG16 | 56.8 | 82.3 | - | - | 59.5 |
| BSS Conv Pooling [163] | 2015 | ImageNet [29] | GoogLeNet | - | 88.2 | - | - | - |
| BSS LSTM [163] | 2015 | ImageNet [29] | LSTM [59] | - | 88.6 | - | - | - |
| Two-Stream Fusion [45] | 2016 | ImageNet [29] | VGG16 | 65.4 | 92.5 | - | - | - |
| TSN [144] | 2016 | ImageNet [29] | BN Inception | 68.5 | 94 | - | 19.7 | 73.9 |
| TLE [32] | 2017 | ImageNet [29] | BN Inception | 71.1 | 95.6 | - | - | - |
| I3D [20] | 2017 | ImageNet [29], Kinetics400 [70] | BN Inception | 74.8 | 95.6 | 32.9 | 41.6 | 71.1 |
| ActionVLAD [46] | 2017 | ImageNet [29] | VGG16 | 69.8 | 93.6 | 21 | - | - |

| Method | Year | Pretrain | Backbone | HMDB51 [75] | UCF101 [125] | Charades mAP [121] | Sth-Sth V1 [49] | Kinetics400 [70] |
|---|---|---|---|---|---|---|---|---|
| P3D [107] | 2017 | Sports1M [69] | ResNet50 | - | 88.6 | - | - | 71.6 |
| Non-Local [147] | 2018 | ImageNet [29] | ResNet101 | - | - | - | 44.4 | 77.7 |
| CoViAR [152] | 2018 | - | ResNet50 | 70.2 | 94.9 | 21.9 | - | - |
| R2+1D [137] | 2018 | Kinetics400 [70] | ResNet34 | 74.5 | 96.8 | - | - | 72 |
| TRN [172] | 2018 | ImageNet [29] | BN Inception | - | - | 25.2 | 42 | - |
| S3D [158] | 2018 | ImageNet [29], Kinetics400 [70] | BN Inception | 75.9 | 96.8 | - | 42 | 74.7 |
| Hidden Two-Stream [175] | 2018 | ImageNet [29] | BN Inception | 66.8 | 93.2 | - | - | 72.8 |
| ECO [176] | 2018 | Kinetics400 [70] | BN Inception, ResNet18 | 72.4 | 94.8 | - | 41.4 | 70 |
| TSM [85] | 2019 | ImageNet [29] | ResNet50 | 73.5 | 95.9 | - | 47.2 | 74.1 |
| SlowFast [43] | 2019 | Kinetics600 [18] | ResNet | - | - | 45.2 | - | 79.8 |
| TEA [82] | 2020 | ImageNet [29] | ResNet50 | 73.3 | 96.9 | - | 51.9 | 76.1 |
| X3D [42] | 2020 | - | ResNet | - | - | - | - | 80.4 |
| AssembleNet++ [111] | 2020 | - | ResNet | - | - | 59.9 | - | - |
| MVFNet [153] | 2021 | ImageNet [29] | ResNet101 | 75.7 | 96.6 | - | 54 | 79.1 |
| SDA-TSM [130] | 2021 | ImageNet [29] | ResNet50 | - | - | - | 54.8 | - |
| ViViT [5] | 2021 | ImageNet [29] | ViT [35], BERT [30] | - | - | - | - | 84.8 |
| MViT [38] | 2021 | - | ViT [35], SlowFast [43] | - | - | 47.7 | - | 81.2 |
| STAM [119] | 2021 | ImageNet [29], Kinetics400 [70] | ViT [35] | - | 97 | 39.7 | - | 80.5 |
| TokenShift [166] | 2021 | ImageNet [29], Kinetics400 [70] | ResNet50 | - | 96.8 | - | - | 80.4 |

## 7 OPEN CHALLENGES AND FUTURE WORKS DISCUSSION

Despite the excellent performance in activity recognition and video understanding, spatiotemporal deep neural networks are still far from perfect. In this part, we will provide an overview of some of the open challenges and limitations of the video understanding models.

### 7.1 High Computational Cost

Even with the development of new models aimed at reducing computational costs, feed-forward neural networks in video understanding can still become excessively overparameterized. When these models are trained on short video clips, they are exposed to multiple frames' worth of information as input. This information is then propagated through multiple layers for feature extraction and fusion. In parallel, vision transformer models can exacerbate the issue by introducing even more parameters, leading to heightened computational demands. An empirical study on vision transformer models delved into their scalability, exploring models with up to two billion parameters [164]. This situation underscores the ongoing challenge of managing computational costs and model scalability in the realm of video understanding.

Research has demonstrated that incorporating a substantial number of parameters into transformer networks can enable them to effectively train on larger datasets, even when such datasets might otherwise be underutilized [164]. However, it's important to note that enhancing performance through increased parameterization comes at a significant computational cost. This has prompted a burgeoning area of research focused on finding ways to scale up these models while simultaneously mitigating their computational demands, thus enabling their use in more computationally efficient settings. Balancing model performance and computational efficiency remains a critical challenge in the development and deployment of transformer-based networks for various tasks, including video understanding.

Numerous techniques have been explored to alleviate the computational burden in vision neural networks, such as the use of shifting methods [39,85]. In the context of transformer models, recent advancements have focused on enhancing self-attention mechanisms to reduce computational costs. These include approaches like downsampling [38,146,154], local window-based attention [91,139], axial attention [34,58], low-rank projection attention [133,145,159], kernelizable attention [25,103], and similarity clustering-based attention [73,134]. However, it's worth noting that, despite these efforts, many state-of-the-art vision models remain non-scalable due to their substantial computational requirements. In practice, most of these introduced methods entail a trade-off between model accuracy and computational complexity, making it a continuing challenge to strike the right balance between performance and efficiency in video understanding and other vision-related tasks.

## 7.2 Large Datasets

The task of video understanding in neural networks necessitates the mapping of both spatial and temporal features. These models are designed to extract and fuse the intricate relationships between these spatiotemporal features. Achieving this goal typically demands a substantial amount of training data and supervised learning, as modeling these complex relationships is a challenging task

In contrast, transformer networks lack the inherent capability to encode prior knowledge for handling visual data and often necessitate large datasets to learn the underlying relationships and rules. CNN-based models, thanks to their structural design and pooling operations, come with built-in features like translation invariance, weight sharing, and partial scale invariance. On the other hand, vision transformer models must learn these relationships and types of invariance by processing more input data. For example, models like ViT [35] require training on hundreds of millions of images to achieve reasonable performance. Fortunately, larger video-understanding datasets have been introduced in recent years, providing valuable resources for benchmarking these tasks.

## 7.3 Invariant Feature Detection with Videos

One of the primary challenges in all recognition tasks is the input variance problem. This issue arises when the neural networks designed for object and action recognition become overly sensitive to parameters that are not crucial for the main inference. Consequently, the model struggles to respond accurately to minor changes that may occur in the input data. Some of these changes include alterations in camera angles, variations in brightness, changes in scale, rotations, other transformations, shifts in color, the introduction of obstacles, and the presentation of unrelated frames within a video sequence. While humans can easily recognize objects and actions under such changes in visual input, computer models often struggle. Even slight amounts of noise can significantly impact the results produced by neural networks. The primary reason for this problem lies in the nature of neural networks and their reliance on supervised training, which optimizes the network based on the data it was trained on. In other words, the trained neural networks may lack the necessary generalization to handle the wide array of input variations encountered in real-world scenarios.

While CNN-based models do exhibit partial invariance to transformations and scaling, many video understanding models still struggle with achieving invariant spatiotemporal feature detection. This challenge primarily arises due to the extensive supervised training required on large datasets to enable models to capture these complex relationships. To address this problem, some studies have introduced novel approaches to enhance deep neural networks and make them more robust to variations in input data. For example, NetVLAD has improved the VLAD aggregation method to help mitigate this issue. (Figure 25) illustrates how NetVLAD demonstrates invariance in a place recognition task [2].

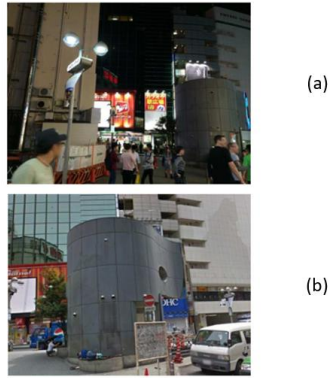

Figure 25. NetVLAD can classify correctly with variant inputs. The images are from the same location. However, the perspective and daylight are changed. Also, multiple obstacles cover the building in the query image. Image a was a mobile phone query and image b was a retrieved image from the database. Figure from [2]

Utilizing videos as input and transitioning towards multi-channel networks offers a promising approach to enhance the robustness and invariance of our predictions. This approach is particularly effective because the key features extracted from videos are inherently spatiotemporal. These spatiotemporal features capture how spatial features change over time. This temporal information enables the network to develop a more generalized understanding of objects and actions, learning to perceive them from various perspectives and under diverse conditions during training. Consequently, the network acquires a more comprehensive representation of these objects or actions, enhancing its ability to recognize them in the future under varying circumstances.

In our recent research [37], we conducted a comparative analysis of several multi-channel architectures to assess their performance in addressing input variance issues. We explored the distinctions between image-based and video-based models concerning their handling of input variance problems.

Highly parametrized networks are often more susceptible to input variance. Nevertheless, implementing mechanisms such as shifting attention and context gating can significantly mitigate the influence of misleading data. Context gating, in particular, proves to be a valuable tool in addressing the invariance problem.

### 7.4 Hardware Efficient Designs and Online Models

Large-scale video models often demand substantial computational power and memory resources. Consequently, they may not perform optimally in resource-constrained environments, such as Internet of Things (IoT) platforms. Developing smaller, more efficient models that can operate effectively on limited hardware resources is essential. This is particularly crucial because many video understanding applications operate in constrained environments, and reducing computational demands not only conserves power but also has environmental benefits.

Moreover, many video understanding applications operate in online environments, where they continuously receive input video streams, either directly from connected hardware or via the internet. To be practical in these scenarios, models must be scalable and capable of processing video input streams continuously. Several methods have been introduced to create versatile structures that support scalable, unidirectional online input streams, enabling efficient video understanding.

Online video understanding faces a significant challenge: limited access to long-term dependencies when processing earlier frames in a video stream. To address this challenge, some models have been introduced to create a long-term dependency bank that is continuously updated. This bank serves as a resource for the model to better understand and interpret the video stream as it progresses [151]. Recursive and feature bank models have demonstrated proficiency in implementing this approach and ensuring access to valuable long-term contextual information.

Another thing to consider is how most video understanding models, for instance, the transformer networks have a rigid structure and are made to work with short snippets and not longer videos. This rigid structure will not allow online processing. Online video streams have a large input rate. Eventually, it can exceed the limits of these models very rapidly. Several simple yet effective methods are employed to preprocess and sample an online stream. Variant types of temporal fusion [69] discussed in earlier sections of this survey can also assist the models with the scalability required for online settings. As for transformer models, [142] reduced the computational cost of the RoBERTa [90] model using hardware-aware transformers (HAT) [9,54,104]. Online and hardware-efficient models will remain one of the most important challenges with video understanding models considering the use cases and mentioned challenges.

## 8  CONCLUSION

In conclusion, this paper has provided an extensive overview of the evolving landscape of video and action understanding within the realm of deep neural networks. We delved into the complexities posed by temporal dimensions, short and long dependencies, and various challenges encountered in video analysis. By examining the extraction and fusion of spatiotemporal features, we highlighted the significant strides made in advancing deep learning techniques for video-related tasks.

The structure of video-based deep neural networks may appear fundamentally similar to that of image-based neural networks. However, due to several unique factors, such as the temporal dimension, the presence of short and long dependencies, and various other challenges, the task of video understanding becomes considerably more complex. Notable issues include the high computational cost, input variance problems, and the inherent disparity between short-term and long-term dependencies.

With the proliferation of video data across various domains, ranging from surveillance and autonomous vehicles to entertainment and healthcare, the ability to extract meaningful insights from these vast datasets has become a critical need. By comprehensively reviewing the latest trends, techniques, and challenges in video analysis, this paper provides a resource for researchers and practitioners aiming to enhance the accuracy, efficiency, and scalability of video understanding models. Moreover, as video-related applications continue to expand, from smart cities to immersive virtual reality experiences, the insights gleaned from this study have the potential to shape the future of AI-driven video comprehension and its applications.